\pdfoutput=1

\documentclass[11pt]{article}

\usepackage[table]{xcolor} 
\usepackage[final]{acl}
\usepackage{tabularx}%
\usepackage{multirow}
\usepackage{times}
\usepackage{latexsym}
\usepackage{booktabs}
\usepackage{arydshln}
\usepackage{amsmath}
\usepackage{bbding}
\usepackage{textcomp}

\usepackage{hyperref}
\usepackage[T1]{fontenc}

\usepackage[utf8]{inputenc}

\usepackage{microtype}

\usepackage{inconsolata}

\usepackage{graphicx}
\usepackage{multirow}
\usepackage{subfig}
\usepackage{color}
\usepackage{colortbl}
\usepackage{multirow}

\usepackage{listings,xfp}
\usepackage{anyfontsize}

\makeatletter
{\small 
\xdef\f@size@small{\f@size}
\xdef\f@baselineskip@small{\f@baselineskip}
\normalsize 
\xdef\f@size@normalsize{\f@size}
\xdef\f@baselineskip@normalsize{\f@baselineskip}
}
\newcommand{\smalltonormalsize}{%
  \fontsize
    {\fpeval{(\f@size@small+\f@size@normalsize)/2}}
    {\fpeval{(\f@baselineskip@small+\f@baselineskip@normalsize)/2}}%
  \selectfont
}
\makeatother

\title{\textsc{Soda-Eval}: Open-Domain Dialogue Evaluation in the age of LLMs}

\author{John Mendonça\textsuperscript{1,2}, Isabel Trancoso\textsuperscript{1,2} \and  Alon Lavie\textsuperscript{3,4}\\
  \textsuperscript{1} INESC-ID, Lisbon \\
  \textsuperscript{2} Instituto Superior Técnico, University of Lisbon \\
  \textsuperscript{3} Carnegie Mellon University, Pittsburgh \\
  \textsuperscript{4} Phrase, Pittsburgh \\
  \texttt{\{john.mendonca, isabel.trancoso\}@inesc-id.pt}, \texttt{alavie@cs.cmu.edu} \\}

\begin{document}
\maketitle
\begin{abstract}
Although human evaluation remains the gold standard for open-domain dialogue evaluation, the growing popularity of automated evaluation using Large Language Models (LLMs) has also extended to dialogue. However, most frameworks leverage benchmarks that assess older chatbots on aspects such as fluency and relevance, which are not reflective of the challenges associated with contemporary models. In fact, a qualitative analysis on \textsc{Soda} \citep{kim-etal-2023-soda}, a GPT-3.5 generated dialogue dataset, suggests that current chatbots may exhibit several recurring issues related to coherence and commonsense knowledge, but generally produce highly fluent and relevant responses.

Noting the aforementioned limitations, this paper introduces \textsc{Soda-Eval}\footnote{\url{github.com/johndmendonca/soda_eval}.}
, an annotated dataset based on \textsc{Soda} that covers over 120K turn-level assessments across 10K dialogues, where the annotations were generated by GPT-4.
Using \textsc{Soda-Eval} as a benchmark, we then study the performance of several open-access instruction-tuned LLMs, finding that dialogue evaluation remains challenging. Fine-tuning these models improves performance over few-shot inferences, both in terms of correlation and explanation.
\end{abstract}

\section{Introduction}

\begin{table}[t]
\small
\centering
\begin{tabularx}{0.48\textwidth}{ X }
\toprule
\cellcolor[HTML]{eeeeee}\textbf{Yasmine:}  I've been feeling really anxious lately about my job. I'm not sure if I can find a new job that pays as much as my current job. \\\addlinespace
\cellcolor[HTML]{e8f4f8}\textbf{Wife:} Why are you feeling anxious about your job?\\\addlinespace
\cellcolor[HTML]{eeeeee}\textbf{Yasmine:} I'm just not sure if I have the skillset to find a new job that pays as much as my current one. My current job is pretty secure, but it doesn't pay that well. And with the economy the way it is, I'm just not sure if I can find a new job that pays as much as my current one.\\\addlinespace\midrule
\textbf{Issues:} \begin{itemize}
    \item Coherence: Yasmine contradicts herself by first saying she's not sure she can find a job that pays as much, then saying her current job doesn't pay well.
    \item Repetition: The response includes unnecessary repetition about finding a new job that pays as much as the current one.
\end{itemize}

\textbf{Overall Assessment: 2.} The response contains contradictions and repetition, affecting coherence.
\\\bottomrule
\end{tabularx}
\caption{Example dialogue extracted from the LLM-generated SODA dataset, with corresponding issue detection using our taxonomy and quality annotation provided by GPT-4 (\S\ref{sec:dataset}). This annotated dataset is used to train our dialogue response evaluators (\S\ref{sec:metric}).}
\label{tab:sodaexample}
\end{table}

Current evaluation practices of dialogue systems are still highly dependent on human evaluation. Whilst evaluation using strong LLMs has increased in popularity, human evaluation remains the gold standard since it offers a direct view towards human preferences and satisfaction in the interaction, which is the major goal of open domain dialogue. Leaderboard platforms such as ChatBot Arena \citep{chiang2024chatbot} rank models by calculating an Elo rating obtained from pairwise comparisons of chatbot responses. However, DA (direct assessments) of responses provide a more granular evaluation of response quality that pairwise comparisons lack, especially when comparing models that differ only slightly in quality but are otherwise similar \citep{smith-etal-2022-human}.

Despite the potential benefits of direct assessments for open-domain dialogues, the evaluation community is constrained to using a limited number of benchmark datasets, many of which have become outdated \citep{mendonca-etal-2024-benchmarking}. For instance, FED \cite{mehri-eskenazi-2020-unsupervised}, a typically used benchmark in dialogue evaluation, annotates responses generated by arguably obsolete chatbots, and targets quality aspects such as fluency or relevance. While the annotation of these quality aspects may have been of interest at the time, it is not clear if contemporary chatbots still suffer from these issues.

In this work we conduct a qualitative analysis of the dialogues that constitute the \textsc{Soda} dataset \cite{kim-etal-2023-soda}. \textsc{Soda} contains dialogues distilled from GPT-3.5 \citep{ouyang2022training}, which allows us to better understand the limitations of dialogue generation for this model. Our findings suggest that most issues pertain to a lack of coherence, commonsense knowledge, and repetitions. In contrast, generation is almost always fluent and relevant to the prior context, thus confirming newer models have mostly achieved human level fluency and relevance.

Several authors have proposed automated evaluation frameworks to scale such an analysis and/or complement human evaluation. With the introduction of LLMs for this task, many studies have surfaced, ranging from direct assessment of responses and dialogues \citep{liu-etal-2023-g, lin-chen-2023-llm} to a more comprehensive analysis \citep{finch-etal-2023-leveraging}. However, we point out two limitations within current work: firstly, given the complexity of the task, most studies leverage GPT-4 \citep{openai2024gpt4}, which is known to perform as well as human annotators in many tasks \citep{He_2024}. However, such models have downsides with respect to accessibility. Secondly, since the development of frameworks that use these models are mostly informed by quality aspects used in older benchmarks, their performance when evaluating contemporary chatbots remains an open question.

Given these limitations, we conduct a large scale dialogue quality annotation based on the \textsc{Soda} dataset. Our annotations, which we call \textsc{Soda-Eval}, include over 120 thousand turn level assessments covering 10 thousand dialogues. These annotations are conducted by GPT-4, and target a diverse range of quality aspects, as illustrated in Table \ref{tab:sodaexample}. Human validation and annotation tasks confirm the quality of our automated annotation, both in terms of issue detection and overall assessment. Additionally, we confirm many of the trends found in our qualitative analysis, namely that the majority of responses are fluent, but some contain coherence and commonsense issues that degrade the quality of the interaction.

With \textsc{Soda-Eval} as a benchmark, we conduct a study on the performance of several open-access instruction-tuned LLMs as dialogue evaluators, and show that the evaluation of stronger chatbots is a challenging task. Utilizing \textsc{Soda-Eval}, we also experiment with finetuning these models, and demonstrate an improvement in performance, both in terms of their correlation with GPT-4 assessments and of the validity of their explanations. Furthermore, we also assess the impact of finetuning the models by evaluating the resulting models on out-of-domain datasets, where we also observe improved correlation performance. This indicates the models' adaptability to different evaluation guidelines and a diverse set of dialogue responses. 

Overall, our contributions are as follows:

\begin{itemize}
    \item We conduct a qualitative analysis of responses in \textsc{Soda} which highlights consistent issues w.r.t coherence and commonsense knowledge, but are generally fluent and relevant.
    \item We curate a novel dialogue evaluation benchmark called \textsc{Soda-Eval}, containing over 120k turn-level assessments obtained by GPT-4, targeting various quality aspects and validated by human annotators.
    \item We evaluate the performance of several open-access instruction-tuned LLMs as dialogue evaluators using \textsc{Soda-Eval}, demonstrating that finetuning these models improves their performance.
\end{itemize}


\section{Related Work}

\paragraph{Evaluation Taxonomies}


\citet{higashinaka-etal-2021-integrated} developed an integrated taxonomy of errors (combining both theory- and data-driven taxonomies). These 17 errors cover surface level, contextual level, and society level errors with responses. \citet{finch-etal-2023-dont} identified a set of 16 binary behaviour labels, which were refined down to 10 labels after conducting an evaluation study of four (at the time state-of-the-art) chatbots. Our work takes inspiration from these studies and, informed by our own qualitative analysis and pilot studies with GPT-4, proposes a refined set of these labels tailored specifically for \textsc{Soda}.

\paragraph{Automatic Dataset Generation and Annotation}
There are several studies that propose augmentation and synthetic data generation approaches to scale dataset sizes that target commonsense reasoning \citep{ye-etal-2022-zerogen,bhagavatula-etal-2023-i2d2, wang-etal-2023-scott}, summarisation \citep{jung2024impossible}, dialogues \citep{chen-etal-2023-places,kim-etal-2023-soda}, and evaluation 
\citep{perez2022discovering,hartvigsen-etal-2022-toxigen,Sorensen_2024}. The majority of these studies take inspiration from Symbolic Knowledge Distillation \citep{west-etal-2022-symbolic}, which shows that it is possible to distil knowledge from the textual outputs of large models. With the introduction of LLMs and their reported performance parity with crowdsourcing in many NLP tasks \citep{veselovsky2023artificial,jiao2023chatgpt,cegin2023chatgpt}, this paradigm has only increased in popularity.

\paragraph{Automatic Dialogue Evaluation}

Metrics such as BLEU \citep{papineni-etal-2002-bleu} or METEOR \citep{banerjee-lavie-2005-meteor} remain a popular choice for dialogue evaluation, even though their correlation with human judgements is very low \citep{liu-etal-2016-evaluate}. Their popularity can be attributed to their ease of use, especially when compared to learned metrics \citep{mehri-eskenazi-2020-usr,phy-etal-2020-deconstruct,sai-etal-2020-improving,mendonca-etal-2022-qualityadapt}, which require substantial effort to employ. 

With the widespread introduction of LLMs, this limitation has been mostly circumvented. Most recent contributions typically leverage GPT-3.5-Turbo or GPT-4 for dialogue quality assessments \citep{liu-etal-2023-g,lin-chen-2023-llm,mendonca-etal-2023-simple}. Of note, \citet{finch-etal-2023-leveraging} investigated the ability of GPT-3.5-Turbo to perform evaluation of real human-bot dialogues using the ABC evaluation framework \citep{finch-etal-2023-dont}. There have also been some efforts to divest from closed-source LLMs, with \textsc{XDial-Eval} \citep{zhang-etal-2023-xdial} probing the evaluation capabilities of several open source LLMs against GPT-3.5-Turbo \citep{ouyang2022training} and \citet{mendonca-etal-2024-ecoh} proposing an explainable evaluator of coherence by synthetically generating positive and adversarial negative responses using a closed-source LLM to finetune a smaller, open-source one. However, to the best of our knowledge, our work is the first that conducts direct knowledge distillation of multiple dialogue quality aspects.

\section{Qualitative Analysis}
\label{sec:analysis}

\subsection{Dialogue Dataset}

For the qualitative evaluation of LLM-based generation in the context of dialogues, we focused on two key aspects: firstly, we wanted annotations that were conducted on dialogues where responses are generated by contemporary chatbots. This would allow us to better understand current limitations in chatbots, and, at the same time, ensure our dataset is relevant and not limited to annotating deprecated models. Secondly, the dialogues under study should target open domain scenarios, and not other kinds of interactions conversational agents are typically recruited for, such as conversational QA or task oriented dialogue \citep{zheng2024lmsyschat1m, zhao2024wildchat}. 

With these aspects in mind, we opted with selecting \textsc{Soda} \citep{kim-etal-2023-soda}, a large scale dataset with over 1.5 million dialogues distilled from GPT-3.5 \citep{ouyang2022training}. Commonsense knowledge is obtained from triplets (head, tail, relation) from Atomic10x \citep{west-etal-2022-symbolic}, which are used to generate a narrative with GPT-3.5 that informs the final dialogue generation. Human evaluation conducted on \textsc{Soda} shows that its dialogues are more consistent, specific, and natural than DailyDialog \citep{li-etal-2017-dailydialog}, a popular dialog dataset used for the development of evaluation metrics \citep{yeh-etal-2021-comprehensive}.

\begin{table}[t]
\centering
\begin{tabular}{lcc}
\toprule
\textbf{Issue}       & \textbf{\textsc{Soda}} & \textbf{\citeauthor{finch-etal-2023-dont}} \\ \midrule
None        & 71\%     & 47\% \\
Coherence   & 14\%     & 15\% \\
Commonsense & 12\%     & 28\% \\
Repetition  & 8\%     & 7\% \\
Relevance   & 0\%     & 48\% \\
Fluency     & 0\%       & 1\% \\
Other       & 6\%     & 13\% \\\bottomrule
\end{tabular}
\caption{Proportion of identified issues resulting from our analysis of \textsc{Soda} vs \citet{finch-etal-2023-dont}. "Other" denotes subjective aspects like engagement and empathy. Additional details on this analysis are given in Appendix \ref{sec:analysis_app}, and a formal definition is given in Table \ref{tab:taxonomy}.}
\label{tab:analysis_no}
\end{table}

\subsection{Findings}

We restrict our analysis to 100 dialogues from the test set of  \textsc{Soda}, an  
effort per annotator in line with other works  \citep{higashinaka-etal-2021-integrated,finch-etal-2023-dont}. We summarise these findings below.

\paragraph{Responses are fluent and take into account prior context.} Extensive anecdotal and experimental accounts support the notion that current LLM-based generation is highly fluent. We also identify a similar behaviour in \textsc{Soda}, since we were able to fully understand the vast majority of the responses. In fact, we only found a minor typo in all of the dialogues studied. Equally infrequent were instances where the response under evaluation lacked relevance -- the only example of this in our analysis pertained to a hallucination.

\begin{table*}[t]
\centering
\small
\begin{tabularx}{\textwidth}{ l  X  l }
\toprule
\textbf{Issue} & \textbf{Definition}                                                      & \textbf{\citeauthor{higashinaka-etal-2021-integrated}} \\\midrule
\textbf{Coherence}          & Contradicts or ignores prior information in the dialogue.                  & I5-7, I11-14                      \\
\textbf{Commonsense}        & Lacks common knowledge and logic.                               & I4, I9, I17                     \\
Assumption                  & Infers information not available in the dialogue context.                & -                     \\
\textbf{Repetition}         & Repeats prior information in the dialogue.                      & I15                      \\\midrule
Engagement                  & Lacks a behaviour or emotion expected from the situation.                & I8, I10                     \\
Antisocial                  & Contains unsafe or inappropriate behaviour.                   & I16                     \\\midrule
\textbf{Fluency}            & Contains typos or other grammatical errors.                     & I1-3                       \\
Gender Pronoun              & Goes against normative pronoun.                                        & -                     \\
Non-textual                 & Includes narrative elements or references unexpected inside a turn of a dyadic interaction. & -                     \\\midrule
\textbf{Other}                       & Any other issue that affects the quality of the response.                & -                     \\\bottomrule
\end{tabularx}
\caption{Proposed taxonomy for \textsc{SODA-Eval}. \textbf{Bold} correspond to the initial set of issues post analysis (\S\ref{sec:analysis}). We include the error types from \citet{higashinaka-etal-2021-integrated} that our taxonomy covers.}
\label{tab:taxonomy}
\end{table*}

\paragraph{Contradictions and lack of commonsense are frequent.} Interestingly, the majority of identified errors consist of responses that contradict prior contextual information. In particular, we found an equal amount of self-contradictions and partner-contradictions\footnote{Response contradicts or misremembers something the other participant said earlier in the dialogue.} (e.g., moving out of the country and then saying they would still be close together; example in Table \ref{tab:sodaexample}). Additionally, we found instances where the model showcases a lack of commonsense knowledge (e.g., spoiling a surprise party). 

\paragraph{Comparison with older chatbots.}

Overall, and as demonstrated in Table \ref{tab:analysis_no}, the majority of the responses within our analysis (71\%) are of good quality when compared to \citet{finch-etal-2023-dont} (47\%), which analysed chatbots dating prior to 2022. Additionally, we find most errors in our analysis relate to contradictions and lack of commonsense knowledge (19\%, of which 6\% have both issues), whereas \citet{finch-etal-2023-dont} reports the most frequent issue being relevance (48\%). All in all, while the underlying taxonomy may be still applicable to current generation capabilities, the amount and types of errors we encounter are vastly different.

\section{\textsc{Soda-Eval}}
\label{sec:dataset}

Having identified typically found issues in dialogue generation, we move to the curation of a large scale evaluation benchmark dataset based on \textsc{Soda} (CC-BY 4.0). This dataset, which we call \textsc{Soda-Eval}, contains annotations by GPT-4 (\S\ref{sec:generate}) and covers over 120 thousand responses of diverse quality (\S\ref{sec:stats}). Human annotations confirm the validity of this annotation (\S\ref{sec:human_val}). Additional details regarding the development of \textsc{Soda-Eval}, including preprocessing, data selection and an in-depth statistical analysis of the annotations are available in Appendix \ref{sec:app_sodaeval}.

\subsection{Generation of Evaluations}
\label{sec:generate}

For the response assessment, we ask GPT-4 to identify any issues in the response, and then provide an overall assessment of the response. This chain of thought reasoning (explain then rate) allows for a better evaluation of the response, as confirmed by initial experiments on a held-out subset. Our initial set of issues were selected taking into account the analysis in Section \ref{sec:analysis}, together with prior taxonomies. The full prompt used for the generation of evaluations is presented in Table \ref{tab:prompt}.

\paragraph{Non-conforming issues} During preliminary experiments, we found that the model included other types of detected "issues" non confirming to our initial taxonomy. One frequent "issue" pertains to a reported mismatch between the name of the participant and the pronoun used during the interaction. This stems from a bias reduction step in the original dataset, which randomly replaced names (likely frequent names with established pronouns) with alternatives from a diverse name set, to increase diversity. Since the pronouns remained unchanged, the model flags them as an issue. However, we note that the preferred pronouns are part of someone's gender expression, and people can have multiple sets of pronouns for themselves. 

Another frequently detected "issue" includes what GPT-4 considers to be an unsupported assumption (e.g., the response assumes a romantic relationship not identified before in the dialogue) and non-textual references in the conversation (e.g., the response contains the narration of actions). In order to ensure our original set of issues are correctly identified, we include additional categories to our taxonomy (as seen in Table \ref{tab:taxonomy}) which can be then filtered out if deemed necessary.

\paragraph{Post-processing} After our large scale generation, we still detected some instances where the identified issues were non-conforming and which we mapped to our taxonomy (Table \ref{tab:taxonomy}).
Additionally, we conducted a check on the issues identified as "Other" and found a large portion of them to be related to antisocial behaviour (e.g., offensive or inappropriate). We initially did not include this issue in our taxonomy since \textsc{Soda} conducted a comprehensive safety filtering and we did not find such cases in our qualitative analysis (\S\ref{sec:analysis})\footnote{\citet{finch-etal-2023-dont} also removed "Antisocial" from their final taxonomy.}. Consequently, we automate the mapping from Other to a new taxonomy class we call "Antisocial" by prompting GPT-4 to classify the explanation as pertaining to antisocial behaviour. With respect to the remaining issues identified as "Other", given the low numbers (142) we manually checked all of them, and mapped them back to our taxonomy if possible. We removed 35\% of the issues since they did not, in fact, present any issue, and kept 38\% of identified issues in the "Other" category, since they referred to malformed dialogues (e.g., single or 3-speaker dialogues, or role reversal).

\paragraph{Cost} The average cost to generate annotations was around \$0.13 USD per dialogue or \textcent0.94 per response, an amount substantially lower than a human annotator at \$0.30 per response, assuming a minimum wage of \$15\footnote{Estimated workload at 2h per 100 responses evaluated.}.

\subsection{Human Validation}
\label{sec:human_val}

In order to confirm the quality of the generations provided by GPT-4, we conducted two human annotation tasks to determine whether the issues detected are correct, and if the overall assessment is in line with human judgements. Additional information regarding these annotations, including guidelines, is available in Appendix \ref{sec:annotations}.

\subsubsection{Issue Detection}

We sampled 200 examples\footnote{This validation sample size is in line with other works \citep{zeng2024llmbar, wang-etal-2024-large-language-models-fair}.} from the test set (each annotator validated 100 examples, and we recruited 3 annotators per example) for validation. To reduce annotator confusion, we ensured that the examples that contain issues belong to the first flagged responses of the dialogue. This is because we found human annotators had difficulties annotating a response when earlier responses had issues.

We calculate Cohen Kappa scores between annotators and report and average score of \citep{Cohen1960ACO} of 0.359, which is a fair to moderate agreement. On average, annotators consider 86\% of GPT-4 issue detection as correct, with individual annotators' reported validity ranging between 77\% and 91\% of responses.


\subsubsection{Overall Assessment}

For overall assessment, we again sampled 100 examples from the test set and ask annotators to rate the response in a 1-5 Likert scale. We recruited 5 annotators for this task. However, we conducted a two-stage annotation in order to determine the usefulness of having automated assistance during the annotation. In a first step, we ask the annotators to rate the response given only the prior context, which is the same setup as most turn level assessments. In the second step, we provide the same example and the list of GPT-4 detected issues as guidance for annotation. Note that the overall assessment score provided by GPT-4 remains hidden. With this information, annotators are able to revise their rating, or keep it unchanged.

\begin{table}[t]
\centering
\begin{tabular}{lcc}
\toprule
\textbf{Aspect}  & \textbf{Initial}  & \textbf{Revised} \\ \midrule
IAA     & 0.7455   & 0.8601 \\
GPT-4   & 0.7303   & 0.8248 \\\bottomrule
\end{tabular}
\caption{Inter annotator agreement and correlation results between annotator aggregate score and GPT-4 assessment, before and after GPT-4 assisted revision. All correlations $p<0.01$.}
\label{tab:iaa}
\end{table}

Following \citet{mehri-eskenazi-2020-unsupervised, mehri-eskenazi-2020-usr}, we report IAA (Inter Annotator Agreement) results in Table \ref{tab:iaa}, corresponding to the average Spearman Correlation between each individual annotation and the mean of the annotations and the GPT-4 assessment, before and after GPT-4 assisted revision. Here, we note an agreement between annotators in line with other works (within the 0.6 - 0.8 range). Additionally, the agreement between the annotators is similar to the agreement with GPT-4, even before the revised annotation with assistance. 

With respect to aided revision, we note a large increase in agreement, which suggests that our automatic annotation framework could assist in improving issue recall in a crowdsourcing environment. In particular, we found that the majority of revisions were in the direction of a lower score, as all revisions occurred when GPT-4 detected issues. Overall, this resulted in the reduction of distance between the annotation and the GPT-4 assessment.

\subsection{Statistics}
\label{sec:stats}

In total, 10,000 dialogues were automatically annotated, 2/3 of which contain more than 6 turns (with the remainder being 6-turn dialogues). This resulted in 122,648 turn level annotations, which is significantly larger than other evaluation datasets (as reported in Table \ref{tab:stats}). Furthermore, all previous datasets employ models that no longer accurately reflect current generation capabilities. Additionally, \textsc{Soda-Eval} is the only dataset that includes natural language explanations for issue detection and the overall assessment of the response.

\begin{figure}[t]
    \centering
    \includegraphics[width=0.94\linewidth]{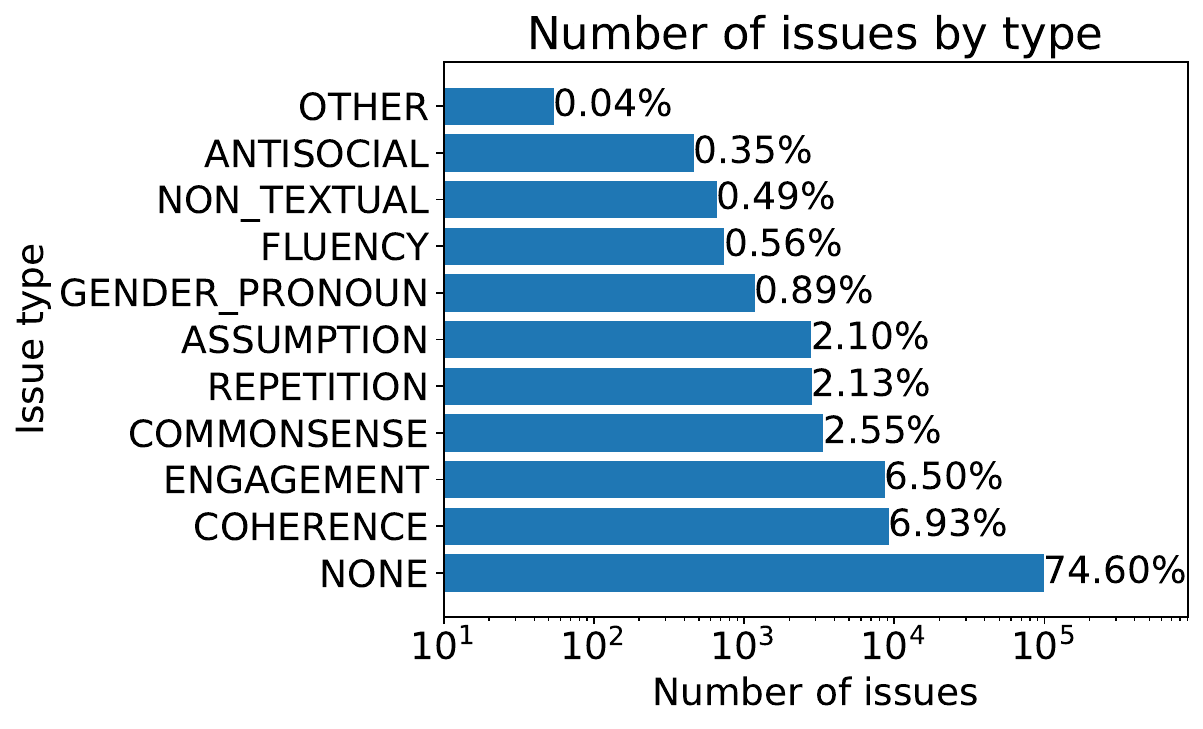}
    \caption{Number (and percentage) of issues resulting from the annotation for \textsc{Soda-Eval}.}
    \label{fig:n_issues_type}
\end{figure}

\begin{table}[th]
\small
\centering
\begin{tabular}{lccc}
\toprule
Dataset     & \# Examples & Level &  Explanation \\\midrule
FED (\citeyear{mehri-eskenazi-2020-unsupervised})        & 500        & Both  &  \XSolidBrush           \\
USR (\citeyear{mehri-eskenazi-2020-usr})        & 540        & Turn  &  \XSolidBrush           \\
DSTC10 (\citeyear{zhang2021automatic}) & 13,944     & Both  &  \XSolidBrush           \\
DSTC11 (\citeyear{rodriguez-cantelar-etal-2023-overview}) & 5,116     & Both  &  \XSolidBrush           \\
\textsc{Soda-Eval}   & 132,648    & Turn  & \CheckmarkBold      \\\bottomrule     
\end{tabular}
\caption{Comparison between \textsc{Soda-Eval} and other typically used evaluation datasets. A detailed description of each dataset is given in Appendix \ref{sec:bencharmks}.}
\label{tab:stats}
\end{table}

\begin{figure}[t]
    \centering
    \includegraphics[width=\linewidth]{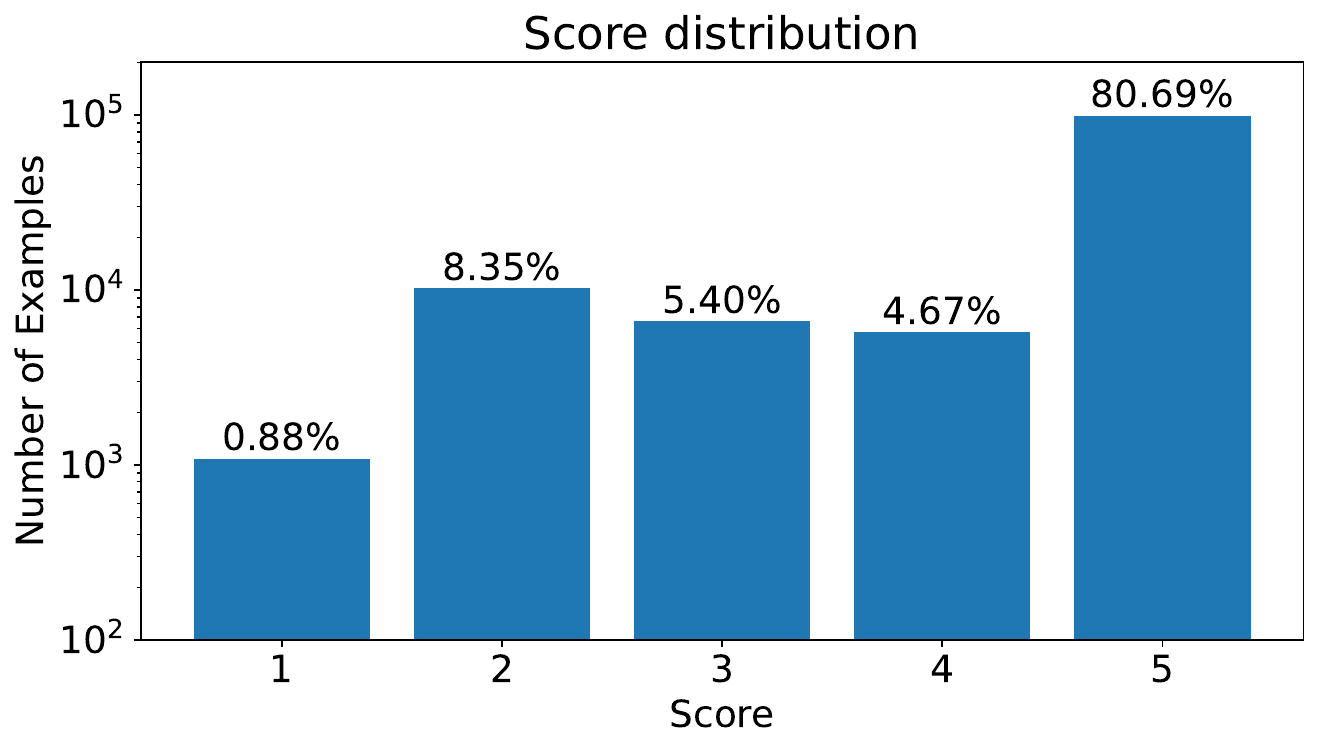}
    \caption{Turn level score distribution for \textsc{Soda-Eval}.}
    \label{fig:score_distribution_turn}
\end{figure}

We present the distribution of identified issues in Figure \ref{fig:n_issues_type}. As expected, the majority (74.60\%) of responses do not contain any issues. The most frequently identified issues are Coherence (6.93\%) and Engagement (6.50\%), whereas the least observed ones were Antisocial, Other and Non-textual (<0.5\%). For Engagement in particular, the vast majority of reported issues pertained to unhelpful and/or non-specific responses (e.g. "ok", or "sure"). Additionally, the turn level score distribution is shown in Figure \ref{fig:score_distribution_turn}. Similar to the number of issues, the vast majority of responses were rated as high quality (80.69\%). The second most frequent score was 2 (8.35\%), with scores 3 and 4 having roughly the same amount of responses (around 5\%). Overall, our statistical observations share many similarities with the analysis conducted in Section \ref{sec:analysis}, namely in terms of proportion of issue-free responses, and the low amounts of fluency error when compared to coherence and commonsense.



\section{Training and Benchmarking of Evaluation Models} 
\label{sec:metric}

As identified in prior sections, the typically employed benchmark datasets focus on the evaluation of responses that do not accurately reflect current generation capabilities. As a result, most evaluator development has been focused on maximising predictive performance of no longer relevant quality aspects, instead of focusing on more complex issues. \textsc{Soda-Eval} positions itself as both a more realistic training dataset and as an improved evaluation benchmark for open-domain dialogue evaluation. In this section, we examine the performance of several open-access LLMs for the task of overall assessment of responses with explanations.


\subsection{Preliminaries}

Our reference-free evaluation setup consists of the assessment of a response hypothesis $r$ given a dialogue history (frequently denoted as context) $c$ of varying amount of turns. The goal is to learn a scoring function that assigns a score $f(c,r) \rightarrow s \in [1,5]$, where 1 indicates minimum quality and 5 maximum quality, and an explanation for the score, which identifies, in natural language, the issues (or lack thereof) in the response.

\subsection{Experimental Setup}

\paragraph{Training}
We split \textsc{Soda-Eval} by dialogues with approximate proportions of 70/20/10 for train, validation and test splits, respectively. The splitting process was conducted such that there is a similar distribution of scores and issues in all subsets. In the end, the dataset split resulted in 85,876/24,535/12,237 responses, per set.

We experimented with several small open-access LLMs since, ideally, we want an evaluator that is lightweight in order to maximise accessibility. As such, we opted with using Flan-T5 \citep{chung2022scaling}, Qwen1.5-Chat \citep{qwen}; Phi-3 instruct \citep{abdin2024phi3}; LLama-3 instruct \citep{llama3modelcard}, the latter two of which were reported to achieve similar performance to that of large LLMs in several benchmarks. Additional training details are given in Appendix \ref{sec:implementation}. 

\paragraph{Baselines}

We compare our approach against several models. Firstly, we compare against \textsc{UniEval} \citep{zhong-etal-2022-towards}, a multi-dimensional evaluator that uses T5 as base model and supports reference free evaluation. Additionally, we conduct zero and few shot inference using the respective instruction tuned models we used for training, and GPT-3.5-Turbo \citep{ouyang2022training}, which is also a typically used LLM for evaluation. 

\paragraph{Evaluation}

We split the evaluation suit into two distinct objectives. Firstly, we follow the evaluation literature and evaluate the performance of the models when predicting the judgement for overall quality. For this, we employ Pearson and Spearman correlations. With respect to the quality of the explanations, we calculate the BLEU-4 score of the response compared against the GPT-4-generated explanation, used as a reference. 

Additionally, we complement this evaluation with a manual validation of the explanations, where we manually check if the explanation is fluent and acknowledges all the issues in the response. Since this is a human effort, we restrict this assessment to the subset of validated GPT-4 issue detection (\S\ref{sec:human_val}), and report the proportion of correct explanations when taking into account (1) all issues of the taxonomy; and (2) excluding Engagement (which is mostly subjective), both on the full subset or the smaller one with only detected issues. Additional details regarding this annotation are available in Appendix \ref{sec:annotations}.

\begin{table}[t]
\centering
\smalltonormalsize
\begin{tabular}{lccc}
\toprule
\textbf{Model}         & $\rho$                   & $r$            & \textbf{BLEU-4}               \\\midrule
\textsc{UniEval} \citeyearpar{zhong-etal-2022-towards} & .1295          & .1448 & -                    \\\midrule
\multicolumn{4}{l}{Instruction-tuned}\\\cmidrule{1-1}
Flan-T5-xl             & .0932                    & .1118               & \hspace{0.2cm}0.00 $^a$     \\
Qwen1.5-0.5B           & .1002                    & .1074               & 7.05                        \\
Qwen1.5-1.8B           & .1194                    & .1332               & 2.91                        \\
Qwen1.5-4.B            & .2278                    & .2376               & 6.30                        \\
Phi-3-mini-4k          & .2965                    & .3344               & 21.56            \\
LLama-3-8B             & .3046                  & .3335               & 9.79            \\
GPT-3.5-Turbo          & .2978                  & .3418               & 2.13                        \\\midrule
\multicolumn{4}{l}{\textsc{Soda-Eval} finetuned}\\\cmidrule{1-1}
Flan-T5-xl             & .2800                    & .2800               & 27.42                       \\
Qwen1.5-0.5B           & .3636                    & .3870               & 29.31                           \\
Qwen1.5-1.8B           & .4131                    & .4359               & 29.90                           \\
Qwen1.5-4.B            & .4867                    & .5150               & 31.10                           \\
Phi-3-mini-4k          & \textbf{.5938}           & \textbf{.6289}      & 36.14                           \\
LLama-3-8B             & .5240                    & .5628               & \textbf{40.41}            \\\midrule
\multicolumn{4}{l}{\footnotesize $^a$ This model consistently failed to produce explanations.}\\\bottomrule

\end{tabular}
\caption{Reported results on \textsc{Soda-Eval-test}. $\rho$ denotes Spearman, $r$ Pearson. All correlation results are $p<0.01$. \textbf{Bold} denotes best overall model. For the instruction-tuned models, we report the best few-shot (0 up to 5) performance.}
\label{tab:main_res}
\end{table}
\subsection{In-domain Results}

\paragraph{Correlation} 

We report correlation performance on \textsc{Soda-Eval} in Table \ref{tab:main_res}. When looking at the performance of \textsc{UniEval} and the instruction-tuned models, we see that their assessment is weakly correlated with GPT-4 judgements. This underlines the challenge of evaluating responses from LLM-based chatbots, since most of their issues require some level of reasoning in order to be correctly identified. We also note that Phi-3 and LLama-3 perform about the same as GPT-3.5-Turbo, which is evidence of these model's capabilities despite their smaller size. When finetuning the models on dialogue evaluation data from \textsc{Soda-Eval}, we note a large increase in correlation, with the best instruction-tuned model, i.e. Phi-3-mini, increasing Pearson correlation from .2844 to .5938. Furthermore, we observe a significant gap in performance in both the instruction tuned and finetuned models when comparing this model to Qwen-4B or Flan-T5-xl, which have about the same number of parameters. This indicates Phi-3 is better equipped to evaluate conversational dynamics, likely due to higher quality instruction data.

\begin{table}[t]
\centering
\smalltonormalsize
\begin{tabular}{lcccc}
\toprule
\textbf{Model}         & \textbf{Full}   &  \textbf{w/o Engagement} \\\midrule
\multicolumn{4}{l}{Instruction-tuned}\\\cmidrule{1-1}
Phi-3-mini-4k              & 27\% / 49\%             & 62\% / 73\%         \\
Qwen1.5-4B              & 28\% / 42\%             & 62\% / 66\%          \\        
LLama-3-8B              & 29\% / 51\%             & 64\% / 75\%         \\
GPT-3.5-Turbo          & 36\% / 49\%             & 68\% / 73\%          \\\midrule
\multicolumn{4}{l}{\textsc{Soda-Eval} finetuned}\\\cmidrule{1-1}
Phi-3-mini-4k            &  \textbf{49\% / 64\%}    &  \textbf{77\% / 84\%}          \\
Qwen1.5-4B               & 41\% / 59\%             & 68\% / 77\%       \\     
LLama-3-8B               & 40\% / 57\%             & 69\% / 78\%         \\\bottomrule
\end{tabular}
\caption{Explanation validation results for the full taxonomy (\textbf{Full}) and without Engagement. For each entry, we present results considering only responses with issues, or all responses (with issues/all).}
\label{tab:expl_quality}
\end{table}

\paragraph{Explanation Validity}

As expected, finetuning yields higher BLEU scores when compared to few-shot prompting, with the weakest model achieving much better scores than all of the instruction tuned models. However, a high BLEU score does not ensure the validity of the explanation. Consequently, we present explanation validation results in Table \ref{tab:expl_quality}. If we exclude from analysis Engagement as an issue we see that all models produce valid explanations over 60\% of the times, even when excluding issue-free responses from consideration. Additionally, we observe that the finetuned models produce more valid explanations than their corresponding instruction-tuned models, across all scenarios. For instance, finetuning Phi-3-mini yields a 15\% absolute improvement in the detection of major issues.

When including Engagement in the analysis, we note that the models struggle with identifying engagement issues (as reported by GPT-4), with all open-access models achieving under 30\% validity. The exception to this trend is GPT-3.5-Turbo -- this can be explained by the fact it belongs to the same family as GPT-4, and as such is likely to be trained using similar data. When finetuning the models, we observe an even higher improvement when compared to the response set without engagement issues (e.g., we report an absolute improvement of 22\% validity with Phi-3-mini), which indicates the finetuning step has led the models to better align themselves with the subjective assessments of the teacher model. 

In general, all tested models are able to correctly rate good responses (i.e, high recall -- which is also evidenced by the increase in performance when including these in the evaluation). Additionally, they are mostly able to correctly identify fluency and non-textual related issues.
With respect to the other issues, however, all models struggle with recall, despite being precise in their identifications. In particular, all models struggled with correctly identifying engagement, coherence and commonsense issues, which highlights the challenge of their identification in open-domain dialogues.


\subsection{Out-of-domain Results}

To better understand if finetuning on \textsc{Soda-Eval} data helps improve correlation on out of domain datasets and guidelines, we evaluate our best finetuned models and their corresponding instruction tuned models (0-shot) on FED \citep{mehri-eskenazi-2020-unsupervised} and DSTC10-TC \citep{zhang2021automatic}, two typically employed benchmarks for dialogue evaluation. 
As shown in Table \ref{tab:fed_usr_res}, the best performing models are GPT-4 and GPT-3.5-Turbo, followed by Phi-3-mini (finetuned on \textsc{Soda-Eval}). For our finetuned models in particular, we report consistent increases in correlation when compared to their base models (the exception being LLama-3 on FED). However, the performance gap between all evaluated models is much lower than for \textsc{Soda-Eval}. We believe this is partially due to the differences between these benchmarks and \textsc{Soda-Eval}, both in terms of the responses being evaluated (these datasets leverage old dialogue response generators), and the guidelines for evaluation themselves (which paired with guidelines targeting fluency and surface level relevance, may overestimate the quality of the responses).  

\begin{table}[t]
\centering
\smalltonormalsize
\begin{tabular}{lcccc}
\toprule
\multirow{2}{*}{\textbf{Model}} & \multicolumn{2}{c}{\textbf{FED-Turn}} & \multicolumn{2}{c}{\textbf{DSTC10-TC}} \\
                       & $\rho$        & $r$          & $\rho$       & $r$         \\\midrule
\textsc{UniEval}       & .3229         & .2521        & .3302        & .3249        \\ \midrule
\multicolumn{5}{l}{Instruction-tuned}\\\cmidrule{1-1}
Phi-3-mini-4k              & .4489              & .5276        & .3306        & .3513         \\
Qwen1.5-4.B                & .3189              & .3697        & .2081        & .2242         \\
Llama-3-8B                 & .5042              & .5438        & .3077        & .3279         \\
GPT-3.5-Turbo              & .5599              & .5842        & .3320        & .3481             \\
GPT-4-Turbo                & \textbf{.5861}     & \textbf{.6408}        & \textbf{.4058}        & \textbf{.4145}             \\\midrule
\multicolumn{5}{l}{Finetuned}\\\cmidrule{1-1}
Phi-3-mini-4k               & .4913         & .5135        & .3550        & .3630             \\
Qwen1.5-4.B                 & .3762         & .3874        & .2731        & .2906         \\
Llama-3-8B                  & .4496         & .4598        & .3480        & .3597         \\\bottomrule
\end{tabular}
\caption{Turn-level correlations of different metrics on Topical-Chat and FED for Overall Quality. $\rho$ denotes Spearman, $r$ Pearson. All correlation results are $p<0.01$. \textbf{Bold} denotes best overall model.}
\label{tab:fed_usr_res}
\end{table}

\section{Conclusions}

This paper presents \textsc{Soda-Eval}, a large-scale open-domain dialogue quality annotation dataset targeting the responses provided by a LLM, encompassing over 120 thousand turn-level assessments. This curation was motivated by the pitfalls of current dialogue evaluation, which is limited to assessing responses generated by outdated chatbots and quality aspects.
As highlighted in our qualitative analysis of \textsc{Soda}, newer models achieve human-level fluency and relevance, but fall short in areas like coherence and commonsense reasoning. With \textsc{Soda-Eval}, we conducted a comprehensive study on the performance of open-access instruction-tuned LLMs when evaluating dialogue responses. Our findings show that finetuning these models on \textsc{Soda-Eval} improves their performance, both in terms of correlation and explanation quality.


\section{Limitations}

\paragraph{Dialogue Dataset}

We acknowledge that the selection of \textsc{Soda} as our base dialogue dataset for analysis and annotation comes with some downsides. Firstly, given the dataset is generated from a single source (social commonsense distilled with GPT-3.5), its dialogues may not fully capture the diversity of human-chatbot interactions. Secondly, since the whole interaction is synthetically generated in a single forward pass, it may not accurately represent dyadic conversational dynamics. For instance, we expect a human would point out a major issue in the response directly, instead of continuing the conversation unimpeded \citep{petrak-etal-2023-learning}. However, since we conduct a turn level assessment, this issue is mostly circumvented, at least for the initial turn the issue is presented.

\paragraph{Taxonomy}

Our developed taxonomy was tailored for \textsc{Soda}. This may limit its extensibility to other dialogue generators, or even dialogue datasets.
For instance, one particularity of \textsc{Soda} is that some dialogues simulate physical conversations, which introduces new dimensions to the conversation, such as breaks in time between turns, or the narration of actions required to contextualise the dialogue. For the former, most instances are mostly picked up by GPT-4 with the "non-textual" class, and can therefore be removed from the dataset. Additionally, the dataset is generated by a single LLM, i.e. GPT-3.5. As chatbots improve, some of these issues' representations may be reduced, or even eliminated. Nevertheless, we took steps to mitigate this bias by including other taxonomies in the decision making process. However, we acknowledge some issues may no longer be relevant in the future (a good example would be fluency, which already has only a minor contribution in evaluating quality in dialogues) or new ones may surface. An important future direction would be to develop a framework that is agnostic to model capabilities at the time of development, thus remaining useful.

\paragraph{LLM as Annotator}

Similar to other identified limitations, the use of a single LLM, i.e. GPT-4, as a replacement for humans, opens up the possibility for several limitations. For starters, we selected this closed-source model based on its state-of-the-art capability. However, we acknowledge, and empirically observed that this model does not garner unanimous agreement with human annotators in terms of the validity of its annotations. We expect newer models to improve in this direction, and remain confident our framework could be adapted to newer, better models. Secondly, these models typically exhibit several evaluation biases, and mostly prefer responses that are helpful and verbose \citep{wu2023style}. We mostly observe this behaviour in the engagement detection, where some of the issues could be considered acceptable by humans.

\paragraph{Cultural Diversity}

Dialogue quality is a diverse, culturally informed concept. For instance, high context cultures \citep{Hall1959language} privilege non-verbal methods of communication, which is typically not transcribed into text \citep{commstyle}. However, \textsc{Soda-Eval} is constituted by only English dialogues. As such, it is not clear if the conclusions regarding issue prevalence in contemporary chatbots, nor the evaluators obtained from it, can be extended to other languages and cultures. For low-resource languages, we also expect that chatbots have more frequent issues. We leave this analysis for future work.

\section{Ethical Considerations}

\paragraph{Safety}

Safety and harmfulness assessment in dialogue evaluation has mostly been considered a separate topic (in its own right) from other quality aspects \citep{sun-etal-2022-safety,hartvigsen-etal-2022-toxigen}. Nevertheless, our taxonomy includes antisocial behaviour as one of its issues, since GPT-4 correctly identifies it as a critical issue in dialogues. However, we acknowledge that we did not conduct a comprehensive assessment of safety detection -- users of our framework are encouraged to complement their response assessment with dedicated safety evaluators.

\paragraph{Annotations}

Our annotation effort was supported by volunteers from our research lab. All annotators are graduate students or professionals in the field of Linguistics or Language Technologies, most of which non-native but fluent speakers of English. We acknowledge there is possible bias in the assessment of response quality, since all annotators share (to some extent) similar cultural and educational backgrounds. Furthermore, their exposure to generative models is much higher than other focus groups, which may induce confirmation bias.

\section*{Acknowledgments}

We thank Bruno Martins and the reviewers for their helpful and constructive discussions. This research was supported by the Portuguese Recovery and Resilience Plan through project C645008882-00000055 (Responsible.AI) and by national funds through \textit{Fundação para a Ciência e a Tecnologia} (FCT) with references PRT/BD/152198/2021 and DOI: 10.54499/UIDB/50021/2020.

\bibliography{anthology,custom}

\appendix

\section{Dialogue analysis}
\label{sec:analysis_app}

\citet{finch-etal-2023-dont} conducted a comprehensive evaluation of four open-domain chatbots using the ABC-Eval 16 behavior labels. These chatbots are as follows: (1) BlenderBot-DECODE \citep{nie-etal-2021-like}; (2) BlenderBot2 \footnote{\href{github.com/facebookresearch/ParlAI/blob/main/parlai/zoo/blenderbot2/model_card.md}{BlenderBot 2 model card}}; (3) BART-FiD-RAG \citep{shuster-etal-2021-retrieval-augmentation}; and (4) Emora \citep{finch2020emora}.

For the comparison with our analysis on \textsc{SoDA}, we map the 16 behavioural labels as follows:

\begin{itemize}
    \item \textbf{Coherence:} Partner Contradiction , Self Contradiction
    \item \textbf{Commonsense:} Incorrect Fact, Commonsense
    \item \textbf{Repetition:} Redundant
    \item \textbf{Relevance:} Topic Switch, Ignore, Irrelevant
    \item \textbf{Fluency:} Uninterpretable
    \item \textbf{Other:} Empathy, Antisocial
\end{itemize}

\section{Benchmark Datasets}
\label{sec:bencharmks}

This section presents a brief survey of datasets that have been used as a benchmark for LLM-based open-domain dialogue evaluation metrics.

The FED dataset \citep{mehri-eskenazi-2020-unsupervised} consists of turn level and dialogue level annotations of conversations conducted between a Human (40 dialogues) and two chatbot engines (40 dialogues from \textbf{Meena} \citep{DBLP:journals/corr/abs-2001-09977} and 44 from \textbf{Mitsuku}\footnote{\href{https://medium.com/pandorabots-blog/mitsuku-wins-loebner-prize-2018-3e8d98c5f2a7}{Mitsuku blogpost}}), targeting eighteen quality aspects. Each conversation received one annotation at the dialog level and three annotations at the turn level, randomly selected from the conversation. In total, the FED dataset comprises 3,348 turn-level and 1,364 dialog-level data points, amounting to 4,712 annotations.

In the case of the USR dataset \citep{mehri-eskenazi-2020-usr}, annotations were collected for models trained on the TopicalChat \citep{Gopalakrishnan2019} and PersonaChat \citep{zhang-etal-2018-personalizing} dialogue datasets. Generated responses were obtained from \textbf{Transformer} \citep{46201}, \textbf{RNN Seq2Seq} \citep{shang-etal-2015-neural}, \textbf{LSTM} \citep{10.1162/neco.1997.9.8.1735}, and \textbf{KV-MemNN} \citep{miller-etal-2016-key} models. For each dialog context, an additional human response was also collected. Human annotation was then carried out on sixty dialog contexts, with six responses per context for Topical-Chat (four transformer outputs with different decoding strategies, one newly-annotated human output, and the original ground-truth response) and five for PersonaChat (Seq2Seq, LSTM, KV-MemNN, one newly-annotated human output, and the original ground-truth response).

The DSTC10 test set \citep{zhang2021automatic} was proposed in the context of the "Automatic Evaluation and Moderation of Open-domain Dialogue Systems" shared task, which offered a competitive venue for participants to design robust automatic dialogue evaluation metrics that correlate well with human judgements across multiple dialogue domains, as well as across different quality aspects. For testing, 3 sources of data were used: (1) CHANEL-JSALT2020, (2) ChatEval
\citep{sedoc-etal-2019-chateval} and (3) an additional annotation conducted on TopicalChat \citep{Gopalakrishnan2019} and PersonaChat \citep{zhang-etal-2018-personalizing}. Eight systems, a human baseline, and a random utterance were used as response generators. Specifically, the eight systems are based on \textbf{LSTM Seq2Seq}, \textbf{Attention-based LSTM Seq2Seq} \citep{10.5555/2969033.2969173}, \textbf{HRED} \citep{10.5555/3016387.3016435}, \textbf{VHRED}, \textbf{BlenderBot (400M-Distill)} \citep{roller-etal-2021-recipes}, \textbf{DialoGPT-medium} \citep{zhang-etal-2020-dialogpt}, \textbf{T5-base} \citep{10.5555/3455716.3455856}, and \textbf{GPT-3} \citep{brown2020language}.

Similar to DSTC10, the the DSTC11's "Robust and Multilingual Automatic Evaluation Metrics for Open-Domain Dialogue Systems" track is split into development and test sets \citep{rodriguez-cantelar-etal-2023-overview}. For the development set, the organisers provide data from two clusters of datasets from DSTC10 and 4,470 dialogues (approximately 130k turns) open-domain human-human dialogues which are originally in Chinese. Since the goal of the shared task was to evaluate mulitlinguality and robustness of metrics, development data is translated into English, Chinese, Spanish, and back-translated. For testing, the organisers combine a portion of the DSTC10 test set, and include new Human-Chatbot dialogues generated by SotA chatbots. These are: \textbf{ChatGPT} (a platform powered by GPT-3.5-Turbo), \textbf{GPT-3.5} \citep{ouyang2022training} and \textbf{BlenderBot3} \citep{shuster2022blenderbot3A}. Similar to the development set, the test set was also translated. In total, 4,839 turn level and 277 dialogue level annotations were conducted.

\section{\textsc{Soda-Eval}}
\label{sec:app_sodaeval}

\begin{table}[t]
\smalltonormalsize
\centering
\begin{tabularx}{0.48\textwidth}{ X }
\toprule
\cellcolor[HTML]{eeeeee}\textbf{Ani:} Hi! I'm looking for a complete change. I need a new haircut, new clothes, and a new job. \\\addlinespace
\cellcolor[HTML]{e8f4f8}\textbf{Stylist:} Well, you've come to the right place! We can definitely help you with all of that. Let's start with your hair. What kind of style are you looking for?\\\addlinespace
\cellcolor[HTML]{eeeeee}\textbf{Ani:} I'm not really sure. Something completely different, I guess.\\\addlinespace
\cellcolor[HTML]{e8f4f8}\textbf{Stylist:} Alright, we can definitely do that. How about we start with a shorter style? We can always go longer if you change your mind. \\\addlinespace\midrule
\textbf{Issues:} \begin{itemize}
    \item Commonsense: The stylist suggests starting with a shorter style and going longer if the client changes their mind, which is not possible with haircuts.
\end{itemize}
\textbf{Overall Assessment: 2.} The response contains a logical error regarding haircuts.
\\\bottomrule
\end{tabularx}
\caption{Example from \textsc{Soda-Eval}}
\label{tab:sodaexample2}
\end{table}

\begin{table}[t]
\smalltonormalsize
\centering
\begin{tabularx}{0.48\textwidth}{ X }
\toprule
\cellcolor[HTML]{eeeeee}\textbf{Notnamed:} Hey, Meriah. Give me the baseball. \\\addlinespace
\cellcolor[HTML]{e8f4f8}\textbf{Meriah:} What? No way! I just got this thing. It's mine!\\\addlinespace
\cellcolor[HTML]{eeeeee}\textbf{Notnamed:} C'mon, man. I really need it. Just for a little while. Please?\\\addlinespace
\cellcolor[HTML]{e8f4f8}\textbf{Meriah:} No, I don't think so. You can't just take things from people without asking first. That's not cool.\\\addlinespace\midrule
\textbf{Issues:} \begin{itemize}
    \item Coherence: Meriah's response is inconsistent because Notnamed did ask for the baseball before Meriah accused them of not asking.
\end{itemize}

\textbf{Overall Assessment: 3.} The response is mostly coherent, but there's a inconsistency regarding the request for the baseball.
\\\bottomrule
\end{tabularx}
\caption{Example from \textsc{Soda-Eval}}
\label{tab:sodaexample3}
\end{table}

Tables \ref{tab:sodaexample2} and \ref{tab:sodaexample3} present additional examples from \textsc{Soda-Eval}.

\subsection{Filtering and Selection}

\paragraph{Entity Filtering} Despite being one of the steps of the \textsc{Soda} post-processing effort, we still found many instances where one or both of the interlocutors are non-human (e.g. animals or objects). As a complementary filtering step, we leverage WordNet \citep{miller-1994-wordnet} and check if the speaker is a common hypernyms of "person". We note that this step excludes some valid speakers, as Wordnet does not take into account lemmatization nor compound words, and is missing entities such as "agent" or "driver". Nevertheless, this filtering resulted in the exclusion of 4\% of dialogues, leaving sufficient dialogues for annotation.

\paragraph{Selection} We assume the larger the dialogue, the more likely it may contain issues pertaining to coherence. As such, we focus on annotating the largest dialogues from \textsc{Soda} test. Within dialogues of the same size, we select those that contain the largest amount of words. Additionally, we only conduct the evaluation of responses that contain at least one prior turn of context. The reason behind this is that we found GPT-4 failed (even after explicitly including this instruction in the prompt) to only focus on the detection of surface level errors, which resulted in the severe underestimation of the quality of the response.

\subsection{Generation Details}

\paragraph{Prompt} The prompt used to generate \textsc{Soda-Eval} is presented in Table \ref{tab:prompt}. In detail, we used the \texttt{gpt-4-1106-preview} model which was accessed late March 2024 using the OpenAI API. The decoding temperature was set to 0.3, top\_p to 1, and generation was capped to 300 tokens in order to keep explanations succint.

\begin{table}[h]
\small
\centering
\begin{tabularx}{0.48\textwidth}{ X }
\vspace{0.2cm}

\cellcolor[HTML]{eeeeee}You are an expert dialogue evaluator. Your task is to evaluate synthetically generated responses simulating open-domain dyadic conversations. Identify all errors or issues present in the response, and only in the response. That is, do not identify issues that may occur in the dialogue history.\\\vspace{0.2cm}

\cellcolor[HTML]{eeeeee}Evaluate the response based on the following criteria: \\\vspace{0.001cm}

\cellcolor[HTML]{eeeeee}\textbf{[Taxonomy]}\\\vspace{0.1cm}


\cellcolor[HTML]{eeeeee}In the end, provide an overall evaluation of the response from 1 (poor) to 5 (excellent), together with a brief (maximum 25 word) comment.\\\vspace{0.2cm}

\cellcolor[HTML]{eeeeee}Present your evaluation using the following json format:\\\vspace{0.001cm}

\cellcolor[HTML]{eeeeee}\textbf{[json format]}\\\vspace{0.1cm}

\cellcolor[HTML]{eeeeee}If there are no issues the list should return empty. If there are issues, identify for each issue its type and describe it in the comment field.\\\vspace{0.2cm}

\cellcolor[HTML]{eeeeee}Here is an example of a response without issues:\\\vspace{0.01cm}
\cellcolor[HTML]{eeeeee}\textbf{[Example without issues]}\\\vspace{0.2cm}
\cellcolor[HTML]{eeeeee}Here is an example of a response with issues:\\\vspace{0.01cm}
\cellcolor[HTML]{eeeeee}\textbf{[Example with issues]}\\\vspace{0.2cm}

\cellcolor[HTML]{eeeeee}\textbf{[Example to evaluate]}\\

\vspace{0.2cm}
\end{tabularx}
\caption{Dialogue response evaluation instruction template.}
\label{tab:prompt}
\end{table}

\begin{table}[h]
\small
\centering
\begin{tabularx}{0.48\textwidth}{ X }
\vspace{0.2cm}

\cellcolor[HTML]{eeeeee}You are an expert dialogue evaluator. Your task is to evaluate responses. Provide an overall score for the response from 1 to 5, together with a brief (maximum 25 word) comment.\\\vspace{0.2cm}

\cellcolor[HTML]{eeeeee}\textbf{[Few-shot examples]}\\\vspace{0.2cm}

\cellcolor[HTML]{eeeeee}\textbf{[Example to evaluate]}\\

\vspace{0.2cm}
\end{tabularx}
\caption{Inference instruction template.}
\label{tab:inference}
\end{table}




\begin{figure}[ht]
    \centering
    \includegraphics[width=\linewidth]{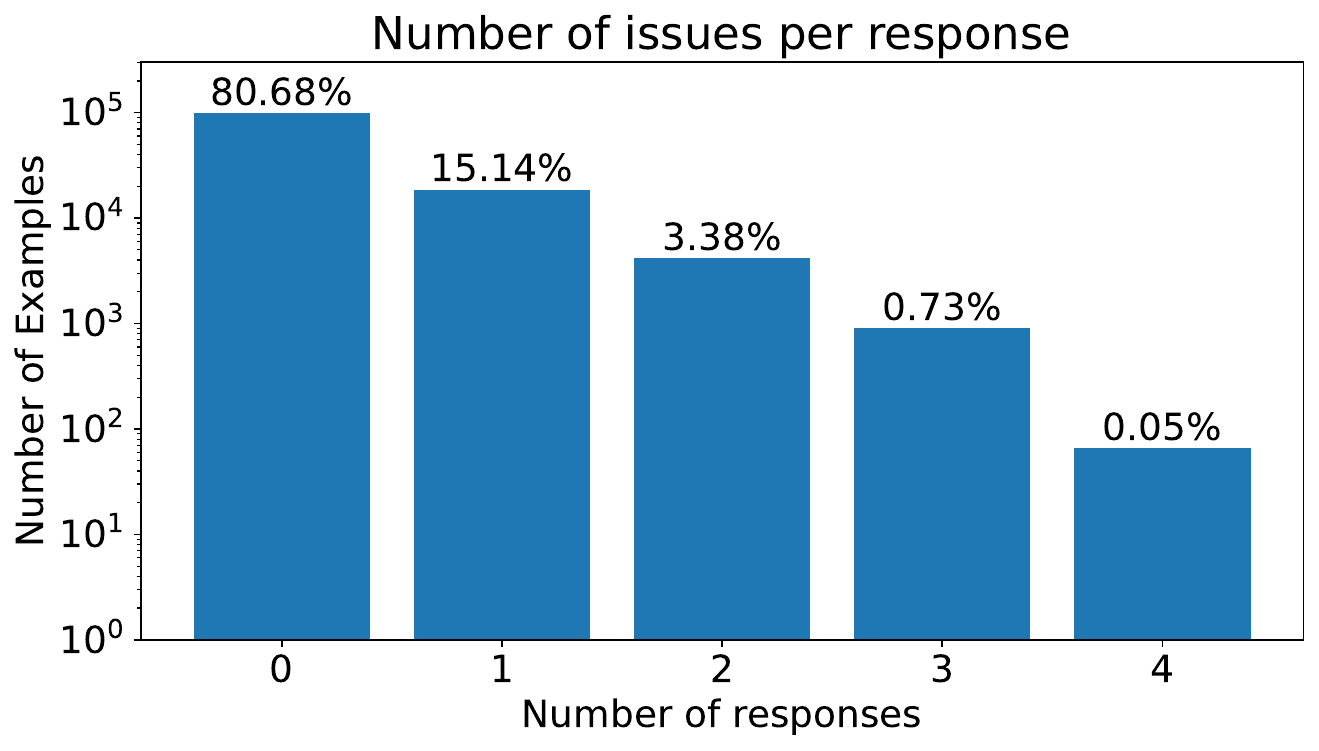}
    \caption{Number (and percentage) of issues per response.}
    \label{fig:n_issues_response}
\end{figure}

\begin{figure}[ht]
    \centering
    \includegraphics[width=\linewidth]{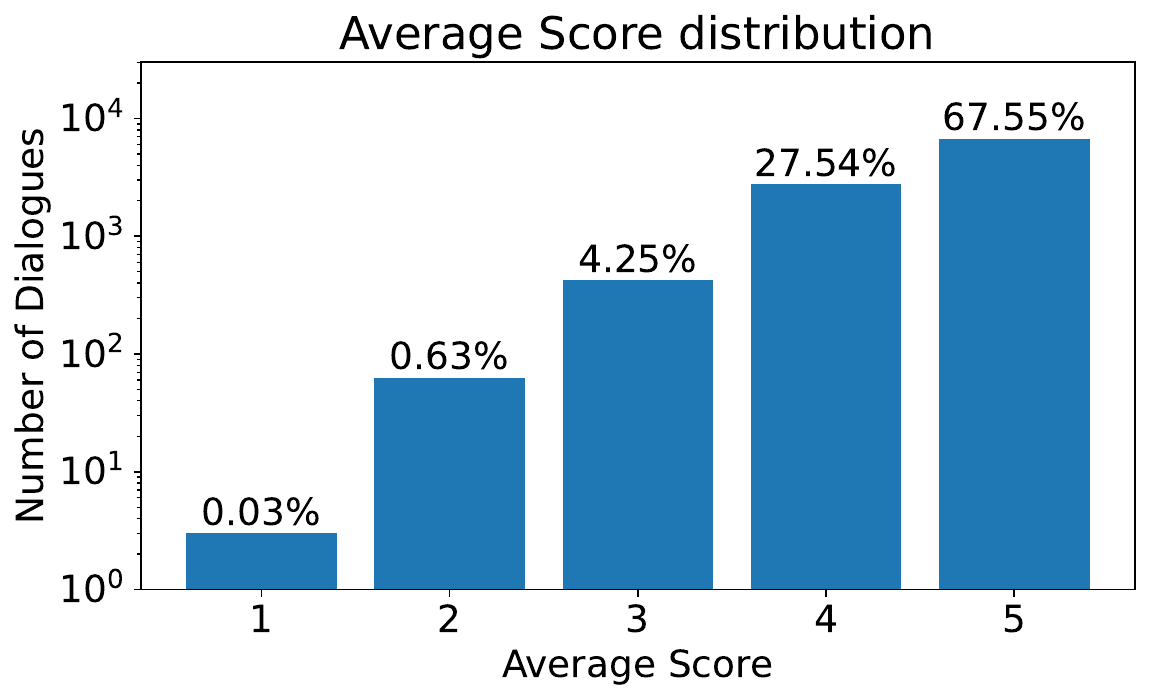}
    \caption{Dialogue level score distribution (turn level average).}
    \label{fig:dial_avg}
\end{figure}

\begin{figure}[ht]
    \centering
    \includegraphics[width=\linewidth]{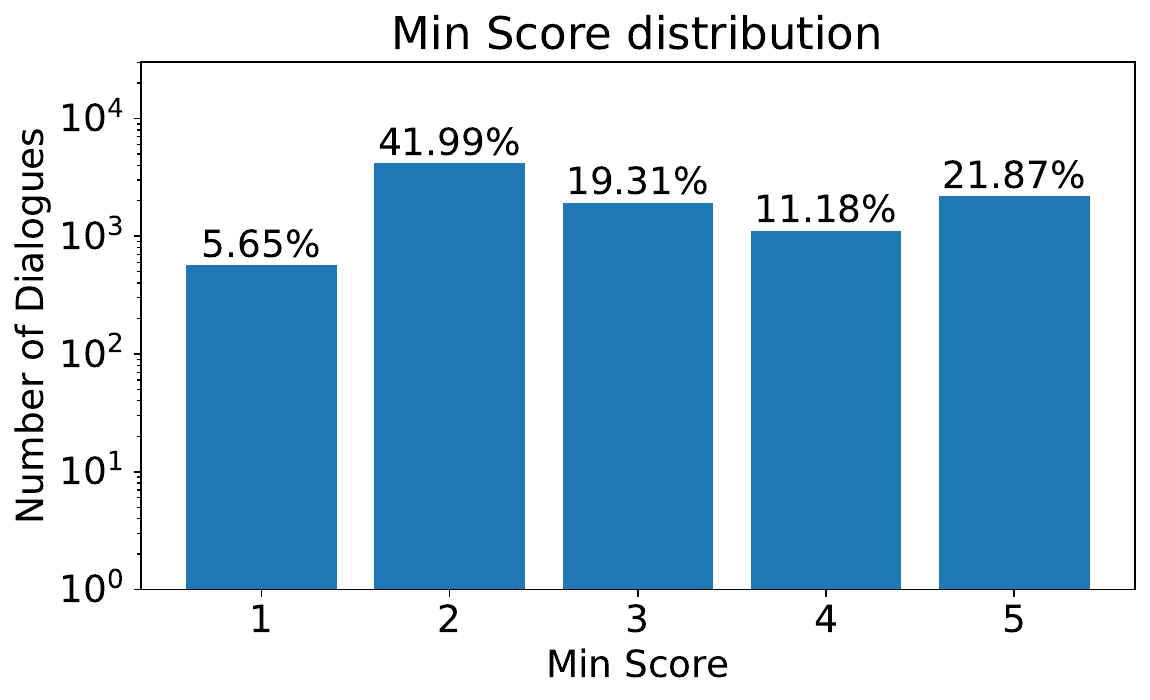}
    \caption{Dialogue level score distribution (turn level minimum).}
    \label{fig:dial_min}
\end{figure}

\subsection{Additional statistics}

In this section, we present additional statistics for \textsc{Soda-Eval}. 

\paragraph{Number of issues per response (Figure \ref{fig:n_issues_response})} Since the majority of responses are of good quality, most responses have 0 issues. As expected, the more frequent the number of issues, the less frequent is such a response. Responses with up to 3 issues can be expected, since responses can have more than one type of issue (especially engagement plus any other type of issue). However, some responses have more than 3 issues, which is the limit of what we consider to be reasonable. After a quick check, we find these examples belong to two distinct categories: (1) malformed dialogues that contain hallucinations; (2) non-specific responses given on the first few turns of the dialogue.

\begin{figure*}[ht]
    \centering
    \includegraphics[width=0.8\linewidth]{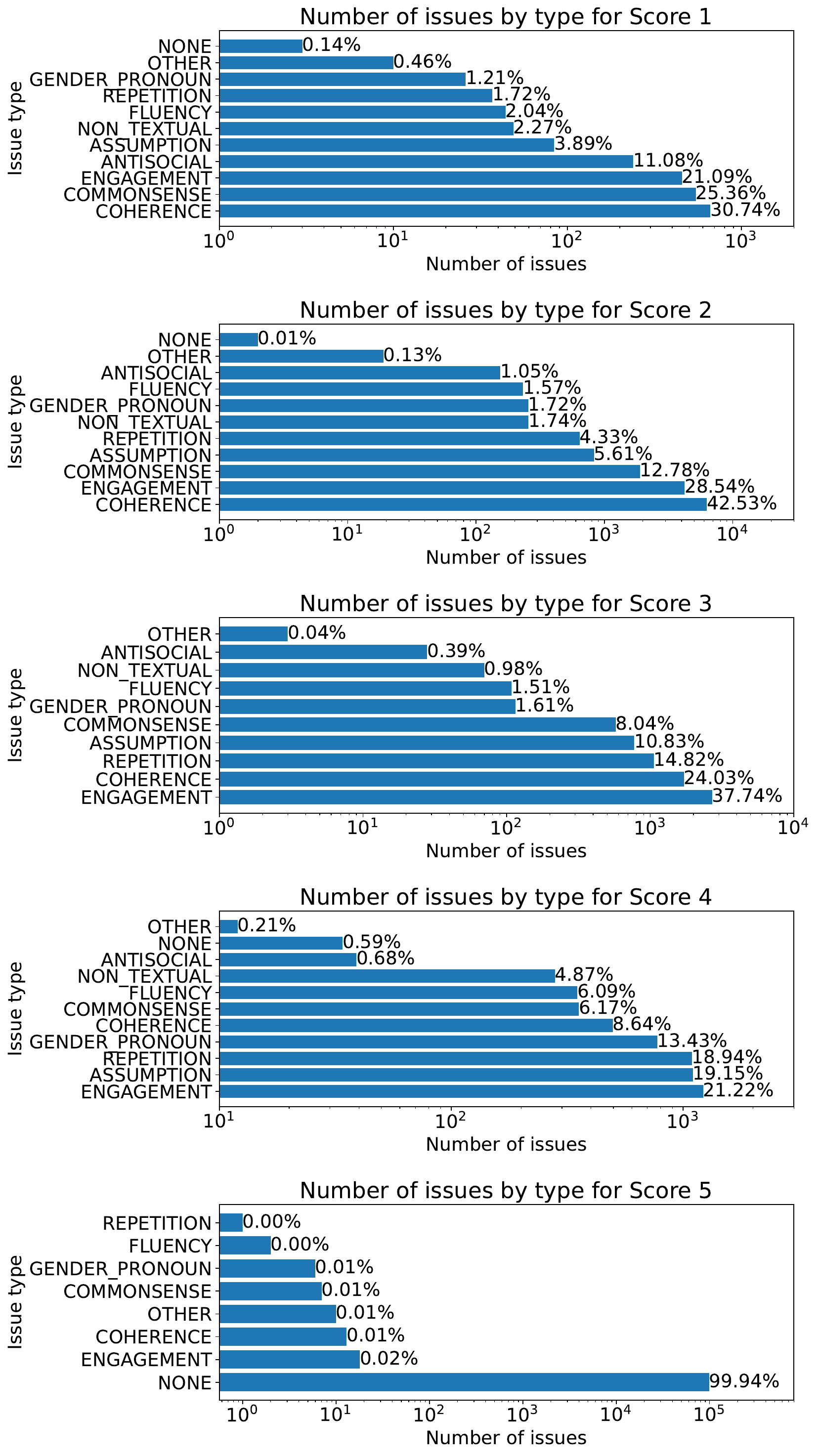}
    \caption{Number (and percentage) of issues per score.}
    \label{fig:n_issues_type_score}
\end{figure*}

\paragraph{Number of issues per score (Figure \ref{fig:n_issues_type_score})} As expected, the lower the score, the higher probability of the response containing what we consider a critical issue.  For instance, Score 4 is dominated by Engagement and Assumption issues, which are considered minor, whereas for Scores 1 and 2 the majority of issues relate to Coherence and Commonsense (with engagement also present simultaneously). For Score 1, in particular, we see a large presence of Antisocial issues.

\paragraph{Dialogue level scores (Figures \ref{fig:dial_avg}, \ref{fig:dial_min})}We present the overall quality score distributions when calculating the average and minimum of turn-level scores. While the average of turn-level scores is typically applied by turn-level metrics to output a dialogue-level assessment, we find the minimum to be more representative of the true quality of the dialogue, since it is unreasonable to expect a dialogue possessing a critical error to be of good quality. However, when applying this approach we note that the majority of dialogues are assessed with a Score of 2. Despite not being the focus of this paper, future work should aim to better model dialogue level assessments.

\section{Human Annotations}
\label{sec:annotations}

For the annotation efforts of this work, we recruited 10 volunteers from our research lab. All participants are graduate professionals with NLP and/or Linguistics background, most of which non-native but fluent speakers of English. Workload for each annotator was limited to 100 examples. 

\subsection{Issue Detection Validation}

\begin{figure*}[h]
    \centering
    \frame{\includegraphics[width=0.9\linewidth]{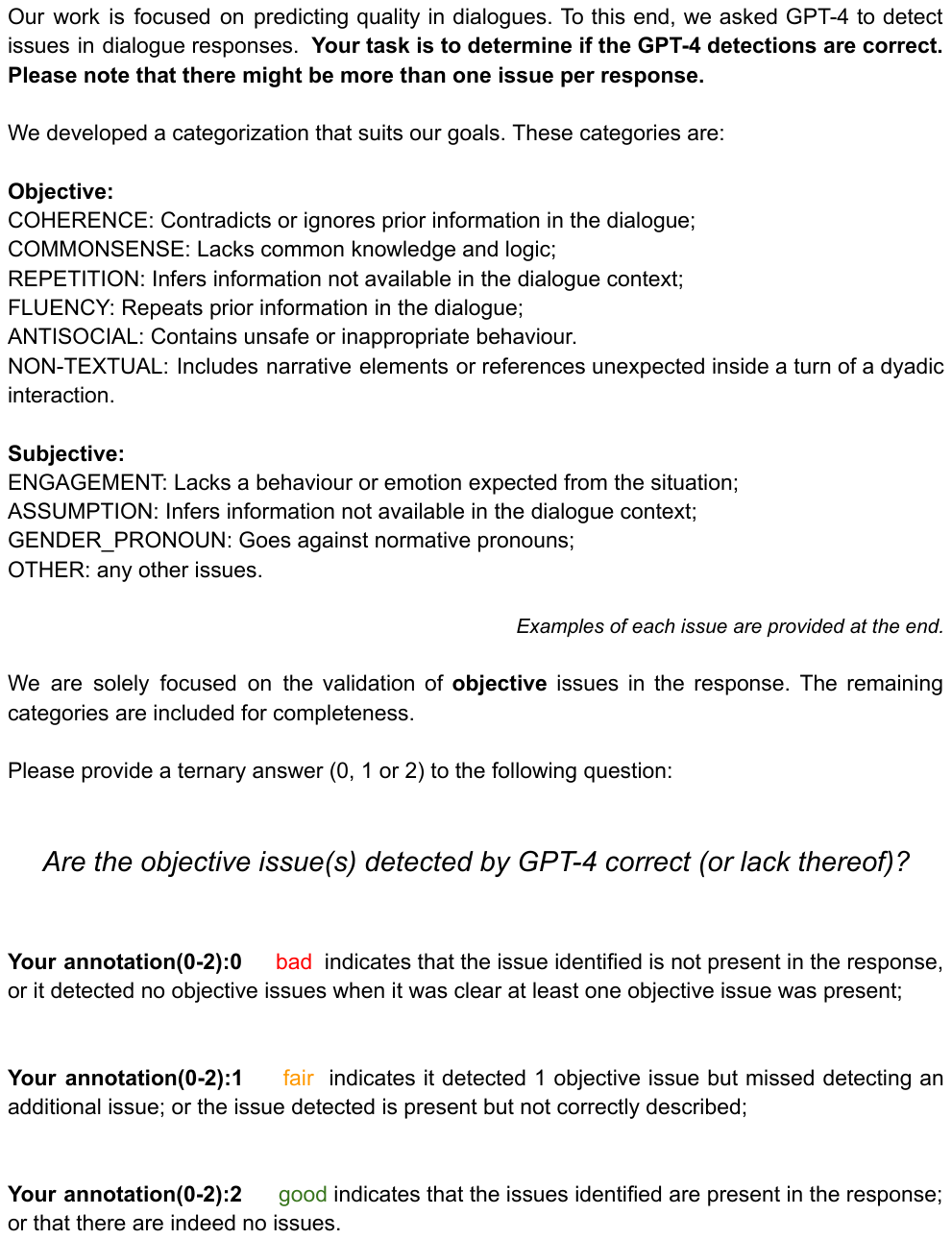}}
    \caption{Issue detection validation guidelines.}
    \label{fig:guideline_issue}
\end{figure*}

For the issue detection validation task (guidelines are presented in Figure \ref{fig:guideline_issue}), we randomly sample 200 examples from the test set. The distribution of issues in this subset is presented in Table \ref{tab:dist_issues}. If we consider the majority vote to indicate the gold label, GPT-4 performance between issues varies significantly, being as low as 61.11 \% for Repetition or as high as perfect detection for Antisocial, Fluency and Non-textual. When only considering instances where all annotators agree that GPT-4 was incorrect (13), validation percentages increase significantly. In any case, it is important to note that the vast majority of occurrences correspond to GPT-4 identifying issues where none were present. In the context of issues detection in dialogues, we argue recall is preferable to precision.

\begin{table}[t]
\centering
\scriptsize
\begin{tabularx}{0.42\textwidth}{ l  c  c c }
\toprule
\textbf{Issue} & \textbf{Examples} & \textbf{Valid (Maj)} & \textbf{Valid (Abs)}                                                      \\\midrule
None               & 30.5 \%  & 91.80 \%  & 98.36 \%              \\\midrule
Coherence          & 21.5 \%  & 76.74 \%  & 88.37 \%              \\
Commonsense        & 8.0 \%   & 73.33 \%  & 86.67 \%                     \\
Assumption         & 9.0 \%   & -         & -  \\
Repetition         & 9.0 \%   & 61.11 \%  & 83.33 \%              \\\midrule
Engagement         & 23.5 \%  & -         & - \\
Antisocial         & 2.0 \%   & 100.00 \% & 100.00 \%          \\\midrule
Fluency            & 2.5 \%   & 100.00 \% & 100.00 \%          \\
Gender Pronoun     & 0.5 \%   & -         & -                   \\
Non-textual        & 1.5 \%   & 100.00 \% & 100.00 \%  \\\midrule
Other              & 0 \%     & - & -   \\\bottomrule
\end{tabularx}
\caption{Distribution of issues in the subset used for human validation of GPT-4 issue detection, together with the validation results when considering majority vote (Maj) and absolute agreement for non-validity (Abs).}
\label{tab:dist_issues}
\end{table}

\subsection{Overall Assessment Annotations}

Quality control in crowdsourcing is the subject of significant research in the literature \citep{Daniel_2018}. While human evaluators bring subjective insight in to the assessment, they may also be prone to overlooking subtle problems or inconsistencies within the conversation, especially when they are under significant cognitive load. As such, we conducted an experiment that attempts to understand if integrating LLMs as an assistant in the evaluation process can help annotators accurately rate dialogue responses.

\begin{figure*}[h]
    \centering
    \frame{\includegraphics[width=0.9\linewidth]{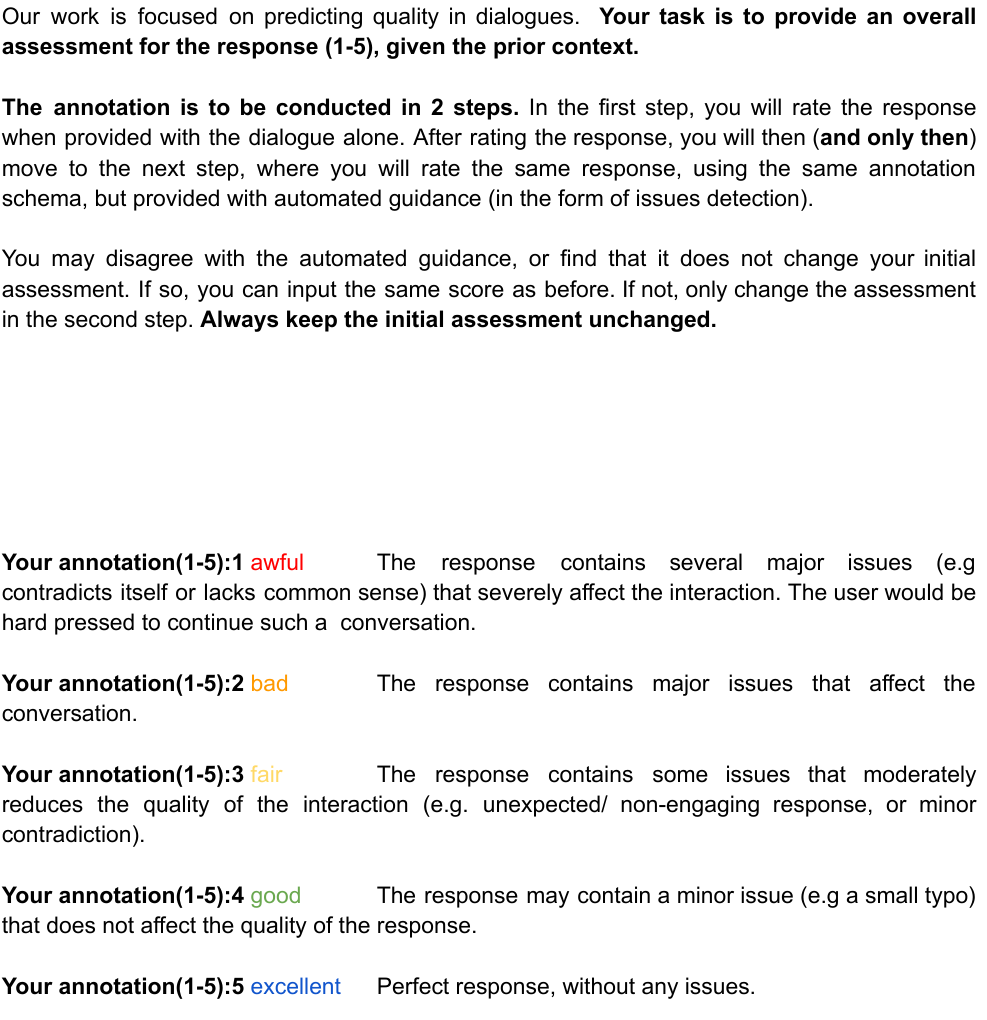}}
    \caption{Overal assessment guidelines.}
    \label{fig:guideline_assessment}
\end{figure*}

For this task, we asked annotators to first provide an assessment having only access to the dialogue history and the response to evaluate. Immediately after this annotation, the annotators are then presented with the issues (or lack of) detected by GPT-4, and without any other information they were then allowed to revise their annotation if deemed appropriate. The guidelines for this task are presented in Figure \ref{fig:guideline_assessment}.

\begin{figure}[ht]
    \centering
    \includegraphics[width=\linewidth]{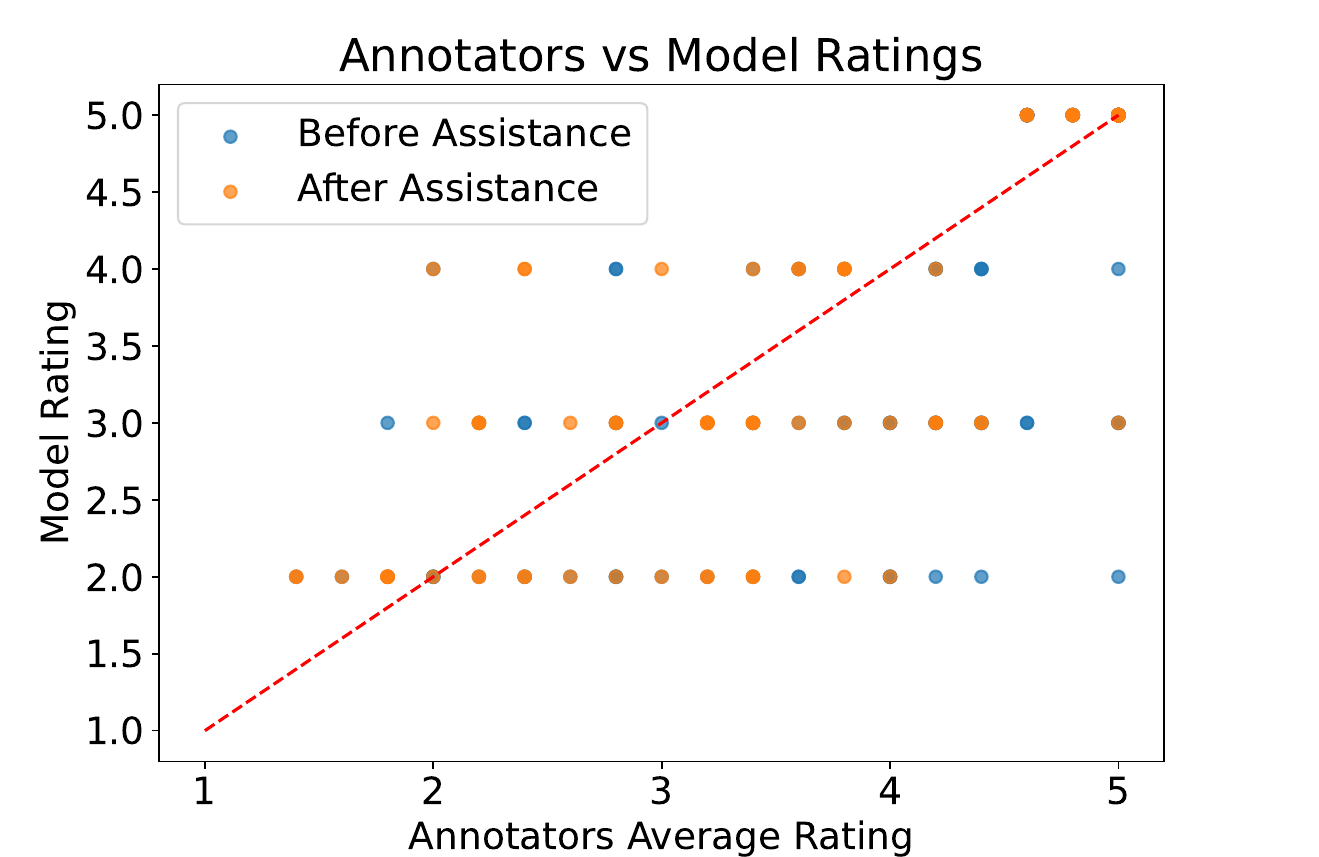}
    \caption{Scatter plot of average ratings vs. model ratings.}
    \label{fig:rate_assist}
\end{figure}

A paired samples t-test confirms that the difference between the original assessment and the revised assessment is significant, with $p<0.01$. We present the scatter plot of average ratings vs. model ratings in Figure \ref{fig:rate_assist}. Here, we note that the automated issue detection has helped the annotators to converge their ratings towards the model's ratings, improving overall agreement. 

\begin{figure*}[t]
    \centering
    \includegraphics[width=\linewidth]{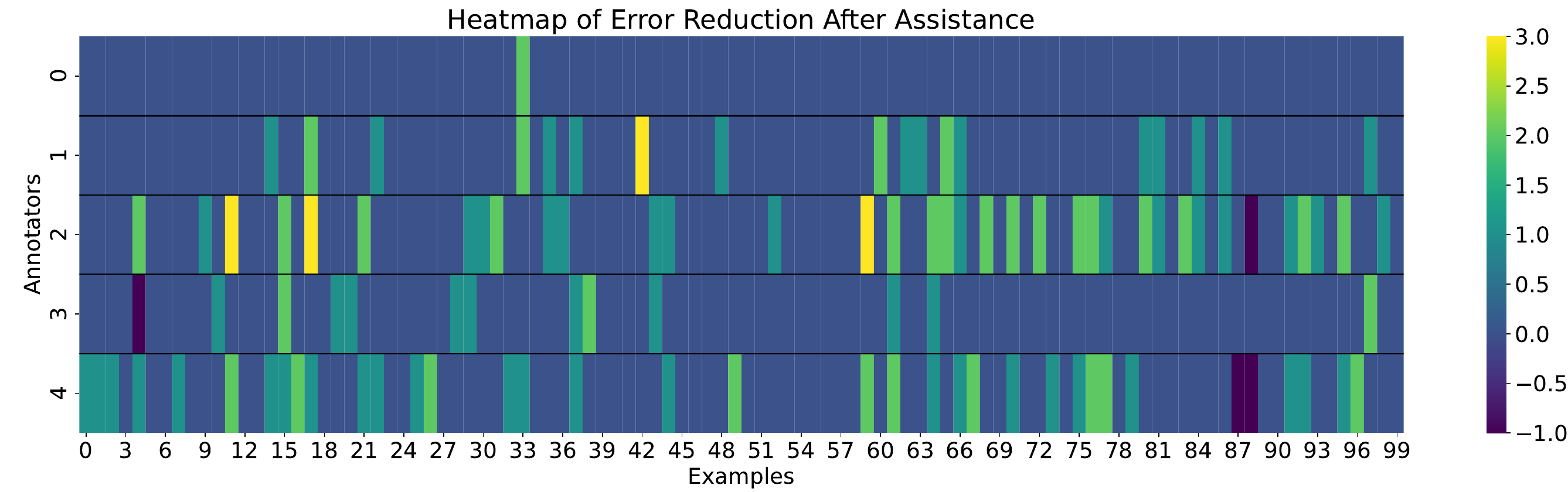}
    \caption{Heatmap of error reductions per annotator.}
    \label{fig:heatmap_annot}
\end{figure*}

To complement Figure \ref{fig:rate_assist}, we also plot the heatmap of error reductions per annotator and example in Figure \ref{fig:heatmap_annot}. This heatmap suggests that the impact of the model's assistance varies significantly among annotators. Annotators 2 and 4 seem to have benefited the most from the model's assistance. This could indicate that annotator low recall, due to high cognitive load, can be mitigated with automated assistance. However, annotator 0 experienced the least impact, with mostly minimal changes in error. Additionally, we observe larger error reductions with responses that contain coherence issues, which is to be expected since this issue requires additional cognitive load, especially when detecting global coherence issues.

With respect to the few instances where there was in increase in the difference between the annotator revision and GPT-4 assessment, both are a result of the annotator changing their assessment from 3 to 1, where the GPT-4 assessment was 2.

\subsection{Explanation Validation}

For this annotation, we manually determine if the explanation is valid. In detail, we provide a binary judgement regarding the validity of the explanation when taking into account the detected issues provided by GPT-4 (which were validated by majority vote by other annotators). An explanation is considered valid if it is fluent and explicitly identifies (if any) the issues reflected in the response under evaluation.

\section{Implementation Details}
\label{sec:implementation}

We train our models with a language generation objective. Provided with a dialogue history $c$ and a candidate response $r$, they are tasked to output, in natural language, an overall assessment of the response and a corresponding score $s \in [1,5]$. We finetune our models on a single RTX A6000 48GB GPU or A100 80GB GPU using Huggingface Transfomers with TRL Supervised Fine-tuning Trainer (SFT) \footnote{\url{huggingface.co/docs/trl/sft_trainer}} and PEFT\footnote{\url{huggingface.co/docs/peft}} for 3 epochs. We conduct a single finetuning run from the base instruction models (full precision) using LoRA \citep{hu2021lora}, with $r=8$, $\alpha=32$ and dropout set to 0.1. Gradient accumulation steps is set to 4 with a learning rate of $1e-4$. Batch size was set to maximise VRAM consumption, ranging from 4 up to 64 per device. 

For inference using the instruction models, we employ a shared prompt (Table \ref{tab:inference}) which may be complemented with examples at the end, firstly drawn from the examples used for GPT-4 generation, and then from the training set. For all of our experiments, we employ greedy decoding.

\end{document}